\documentclass{article}
\usepackage{arxiv}
\usepackage[T1]{fontenc}
\usepackage[utf8]{inputenc}
\usepackage{newtxtext,newtxmath}
\usepackage{microtype}
\usepackage{amsmath}
\usepackage{graphicx}
\usepackage{float}
\usepackage{placeins}
\usepackage{booktabs,tabularx,array}
\usepackage[numbers,sort&compress]{natbib}
\usepackage{xcolor}
\usepackage{tikz}
\usetikzlibrary{arrows.meta,positioning}
\usepackage{enumitem}
\usepackage{hyperref}
\hypersetup{colorlinks=true,linkcolor=blue!45!black,citecolor=blue!45!black,urlcolor=blue!45!black}
\graphicspath{{figures/}}
\setlist[itemize]{leftmargin=*,nosep,topsep=3pt}
\setlist[enumerate]{leftmargin=*,nosep,topsep=3pt}
\newcolumntype{Y}{>{\raggedright\arraybackslash}X}

\definecolor{draftcolor}{RGB}{137,82,14}


\title{ASIWorld: Benchmarking Autonomous Agents\\in a Persistent Survival Sandbox}
\author{%
  Jianfeng Lin$^{1,2}$\thanks{These authors contributed equally to this work.}
  \quad Runze Wang$^{1}$\footnotemark[1]
  \quad Haorong Hong$^{1}$\footnotemark[1]
  \quad Ken Li$^{1,3}$ \\
  Wenbo Hu$^{1,3}$
  \quad Yuxuan Xie$^{1}$
  \quad Zeno Zeng$^{4}$
  \quad Zhongyuan Liu$^{1}$\thanks{Corresponding author.} \\[0.4em]
  \small $^{1}$Lightspeed Studios, Tencent \quad $^{2}$The Hong Kong University of Science and Technology \\
  \small $^{3}$Nanjing University \quad $^{4}$Hunyuan Team, Tencent \\[0.4em]
  \small \href{https://asigameworld.github.io}{\texttt{https://asigameworld.github.io}}
}
\date{}
\renewcommand{\shorttitle}{ASIWorld: A Persistent Survival Sandbox}
\renewcommand{\headeright}{}
\renewcommand{\undertitle}{}
\hypersetup{pdftitle={ASIWorld: Benchmarking Autonomous Agents in a Persistent Survival Sandbox}}
\begin{document}
\maketitle
\begingroup
\setlength{\intextsep}{6pt}
\begin{figure}[H]
\centering\small
\includegraphics[width=\linewidth]{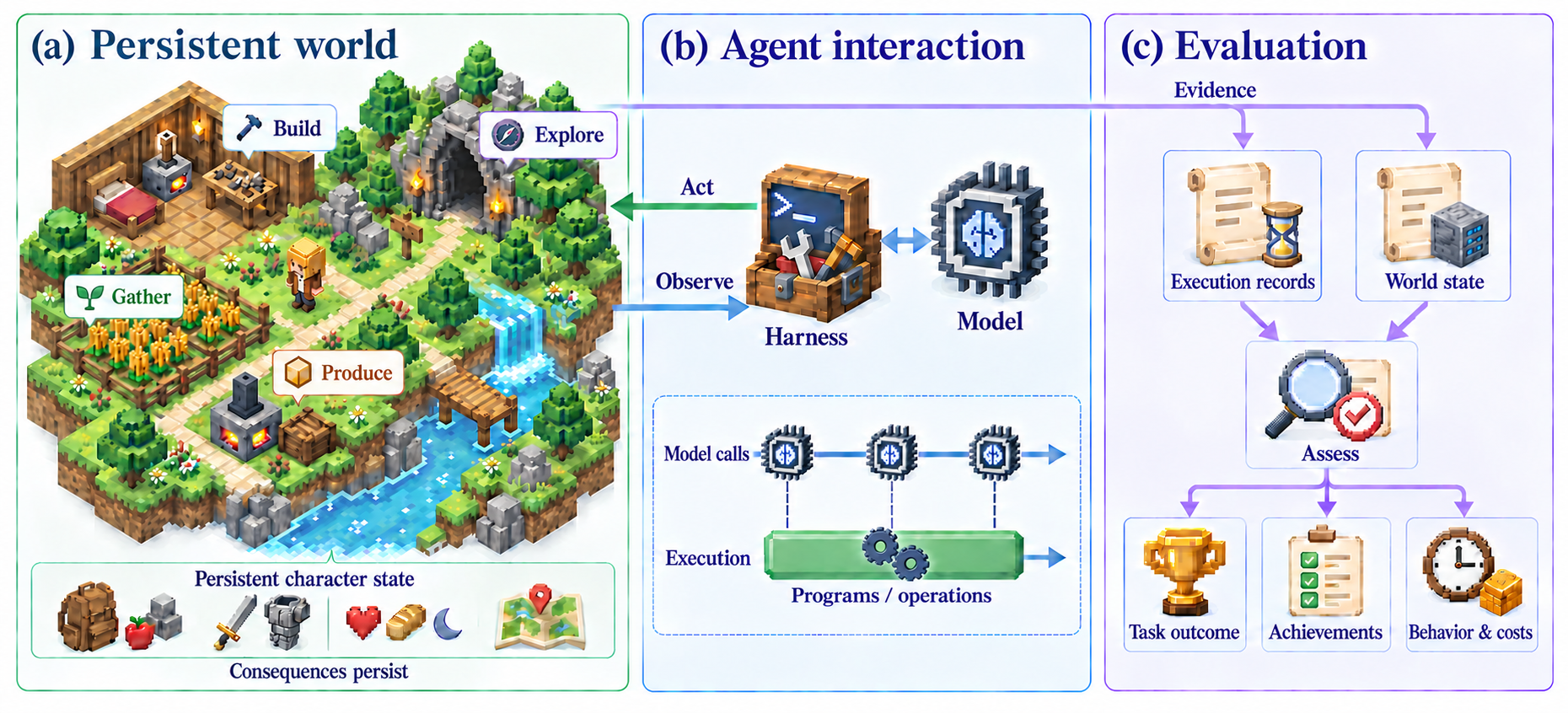}
\caption{\small\textbf{ASIWorld evaluation framework.} (a) Persistent character and world state connects survival, crafting, building, and exploration. (b) Agents interact through a model and execution harness with asynchronous hosted execution. (c) Authoritative world state and execution records support assessment of domain progression, capability diagnostics, behavioral trajectories, and survival and execution costs.}
\label{fig:overview}
\end{figure}
\endgroup
\begin{abstract}
Evaluating autonomous agents requires examining whether they can translate goals into coherent action across extended interactions. In persistent environments, interdependent actions and lasting consequences require assessing both what agents achieve and how they achieve it. We introduce ASIWorld, a persistent open-world survival benchmark for complete agent systems with character-scoped local perception, spatial memory, and direct actions or hosted asynchronous programs. In this persistent survival environment, we benchmark frontier models across survival, combat, building, and crafting to uncover capability disparities in cognition, planning, execution, self-management, error recovery, and social interaction. Our evaluations reveal how models navigate trade-offs between short-term demands and long-term progression across prerequisite planning, spatial reasoning, multi-step execution, and social coordination.

\end{abstract}

\clearpage
\section{Introduction}
Autonomous agents operating in persistent environments must pursue goals across sequences of interdependent actions. With limited observations of a changing world, they must track task state, coordinate these actions, and use execution feedback to guide subsequent decisions. Earlier actions can alter resources, opportunities, and the conditions under which later goals are pursued. Endpoint attainment alone summarizes what an agent ultimately achieves, whereas stage-wise progression, execution behavior, and the associated survival and execution costs provide complementary evidence about how those outcomes are reached. Evaluating sustained goal execution therefore requires relating achieved outcomes to the ongoing process, state maintenance, and operational overhead of execution.

Interactive environments make these demands concrete. Crafter~\citep{hafner2021crafter} and Craftax~\citep{matthews2024craftax} organize evaluation around resource acquisition, survival, and exploration; Voyager~\citep{wang2023voyager} studies sustained skill acquisition in Minecraft; and BALROG~\citep{paglieri2024balrog} evaluates language agents across games. AgentBoard~\citep{ma2024agentboard} complements terminal success with intermediate progress. These perspectives provide a basis for examining goals pursued through successive activities, where preparation and resource expenditure affect later progress. However, prior benchmarks operate in overly idealized settings, evaluating isolated episodes or pre-built action wrappers without coupling a persistent character's execution outcomes with the system-level behavior and survival costs of the agent. We address this gap by studying sustained execution through a persistent character and examining task outcomes alongside the behavior and survival costs of the complete agent system.

We introduce ASIWorld, an AI-native, persistent multiplayer open-world survival role-playing game in which both world and character states continuously change across interactions. Combining open-ended exploration, competitive progression, and social cooperation, the world connects survival, crafting, building, and multi-region combat through this shared, evolving state, where earlier decisions shape the future development of both the character and the world and govern the conditions for subsequent goals. Rather than relying on pre-built game-specific action wrappers or custom plugins, ASIWorld is played directly by a foundation model and general-purpose execution harness through progressively disclosed gameplay guides containing detailed interface documentation, allowing models to inspect relevant interfaces on demand and acquire character-scoped local perception, spatial memory, and a programmable asynchronous interface to control their characters. Agents can issue actions directly or author programs that execute within the environment and continue running between model calls. Feedback from ongoing operations and completed actions informs subsequent decisions and program revisions, creating a setting in which sustained progression depends on coordinating preparation, action, and feedback over time.

Our evaluation treats each agent configuration as a complete system, comprising its model, execution harness, context management, and agent-authored programs. To evaluate how agent capabilities interact with environment demands, we assess systems through a cross-dimensional framework rather than isolated binary endpoints. On the game side, survival, combat, building, and crafting define four interdependent progression domains. On the agent side, cognition, planning, execution, self-management, learning, and social interaction capture six core capabilities for understanding, organizing, and adapting to tasks. Intersecting these two axes grounds each agent capability within each gameplay domain as a verifiable milestone bounded by world ceilings, decoupling overall domain progression from capability-level bottleneck diagnosis. Within this framework, survival costs, including physiological depletion, safe dungeon extraction, and recovery from starvation or combat setbacks, are captured through the survival domain alongside self-management and error-driven learning, while execution costs are measured through total committed actions, retry overhead, and activity allocation across movement, maintenance, and productive domains. Together, these measurements support interpretation of sustained execution while preserving the distinction between observed system behavior and individual underlying capabilities.

In long-horizon evaluations across diverse frontier models using a common harness, we examine how agents progress and evolve distinct behavioral patterns under simultaneous pressures of survival, open-world exploration, and multi-agent competition. While most systems establish reliable basic survival and crafting, they diverge substantially across multi-region combat and face steep bottlenecks in spatial building and socio-economic coordination. Cross-dimensional and trajectory analyses reveal rich behavioral differentiation across autonomous state regulation, closed-loop recovery from execution errors, programmable automation, and dynamic trade-offs between independent advancement and socio-economic cooperation. These observations show how domain progression, emergent behaviors, and execution costs provide complementary accounts of agent performance. We make three contributions:
\begin{enumerate}
\item We build ASIWorld, an AI-native, persistent multiplayer open-world survival RPG spanning survival, multi-tier crafting, housing construction, multi-region combat, and socio-economic interaction, designed to be played directly by a foundation model and general-purpose harness through progressively disclosed gameplay guides, character-scoped observations, and asynchronous hosted programs.
\item We propose a cross-dimensional evaluation framework that intersects the game's four progression domains (survival, combat, building, and crafting) with six core dimensions of agent task understanding and agency (cognition, planning, execution, self-management, learning, and social interaction).
\item We conduct a long-horizon empirical study across frontier model configurations, characterizing how agent capabilities interact with gameplay domains to shape emergent behaviors, error-recovery dynamics, programmable automation, and socio-economic trade-offs.
\end{enumerate}

\section{Related Work}
\paragraph{Environments for long-horizon goal execution.}
Interactive environments expose the dependencies between resource acquisition, preparation, and goal completion. Crafter~\citep{hafner2021crafter} measures crafting, exploration, and survival through achievements, Craftax~\citep{matthews2024craftax} extends these challenges, and MineDojo~\citep{fan2022minedojo} combines Minecraft tasks with multimodal knowledge resources. These settings provide a basis for studying how agents acquire resources and satisfy prerequisites over extended interactions. MineLand~\citep{yu2024mineland} further combines limited perception and physical needs in a multi-agent Minecraft environment. Shared-world research also examines persistent populations in Neural MMO~\citep{suarez2019neuralmmo}, social generalization in Melting Pot~\citep{leibo2021meltingpot}, and long-horizon collaboration in AgentWorld~\citep{shu2026agentworld}. ASIWorld draws on these environment design perspectives to study sustained autonomous progression within a shared survival world. Survival, combat, building, and crafting are connected by persistent inventory, equipment, structures, and physiological state, so preparation and resource expenditure shape the conditions of later actions. Because all evaluated configurations inhabit the same world instance alongside NPCs and other agents, our evaluation examines both individual domain progression and social coordination, ranging from NPC commissions and peer-to-peer trading to cooperative dungeon parties.

\paragraph{Asynchronous interaction and executable behavior.}
Asynchronous environments require agents to coordinate decisions with actions and events that unfold over time. Robotouille~\citep{gonzalezpumariega2025robotouille} evaluates long-horizon planning with overlapping tasks, execution delays, and interruptions, while Gaia2~\citep{froger2026gaia2} studies independently evolving environments with temporal constraints and dynamic events. These works establish asynchronous interaction as an evaluation concern beyond selecting a sequence of actions. A complementary research line examines how model outputs become executable behavior: ReAct~\citep{yao2022react} interleaves reasoning and action, Reflexion~\citep{shinn2023reflexion} incorporates feedback-derived reflections, Voyager~\citep{wang2023voyager} develops executable skills, and Code as Policies~\citep{liang2022codepolicies} represents control logic and perception-based feedback loops as generated programs. ASIWorld brings asynchronous operations and agent-authored programs into a persistent survival setting. Operations can continue after the initiating request, and programs execute within the environment between model calls, while their effects alter resources and character state. This combination connects action scheduling and program execution to subsequent preparation and survival requirements. We therefore evaluate complete agent configurations, including their models and execution harnesses, without attributing observed performance to program execution alone.

\paragraph{Evaluating outcomes, progression, and costs.}
Agent evaluation benefits from distinguishing achieved goals from the behavior and costs associated with reaching them. AgentBench~\citep{liu2023agentbench} covers diverse interactive environments, ScienceWorld~\citep{wang2022scienceworld} studies scientific tasks in a text simulation, and BALROG~\citep{paglieri2024balrog} evaluates language and vision-language agents across games. Beyond task coverage, WebArena~\citep{zhou2023webarena} assesses functional correctness and $\tau$-bench~\citep{yao2024taubench} examines goal-state success and repeated-run reliability. AgentBoard~\citep{ma2024agentboard} analyzes intermediate progress, while AI Agents That Matter~\citep{kapoor2024agents} emphasizes cost-aware evaluation of complete systems. These perspectives motivate complementary measurements rather than a single endpoint score. ASIWorld operationalizes this multi-faceted view by intersecting stage-weighted progression across survival, combat, building, and crafting with diagnostic probes across six core agent capabilities. Alongside domain progression, execution records capture survival costs through physiological maintenance, safe extraction, and error recovery, as well as execution costs through action volume, retry overhead, and activity allocation. Together, these measurements distinguish configurations that reach similar progression levels through different programmatic strategies, activity allocation, and exposure to survival risks.

\section{Environment and Agent Interaction}
ASIWorld is a shared survival environment for studying sustained goal execution, in which each agent controls a character. Characters gather resources, craft equipment, build facilities, and explore a tiered dungeon while meeting their needs for food and rest. Production, construction, exploration, and survival maintenance are connected through a common inventory, equipment, and facilities. World and character states persist across interactions, allowing earlier actions to shape the conditions and opportunities for subsequent tasks.

\begin{figure}[!t]
\centering\small
\includegraphics[width=0.95\linewidth]{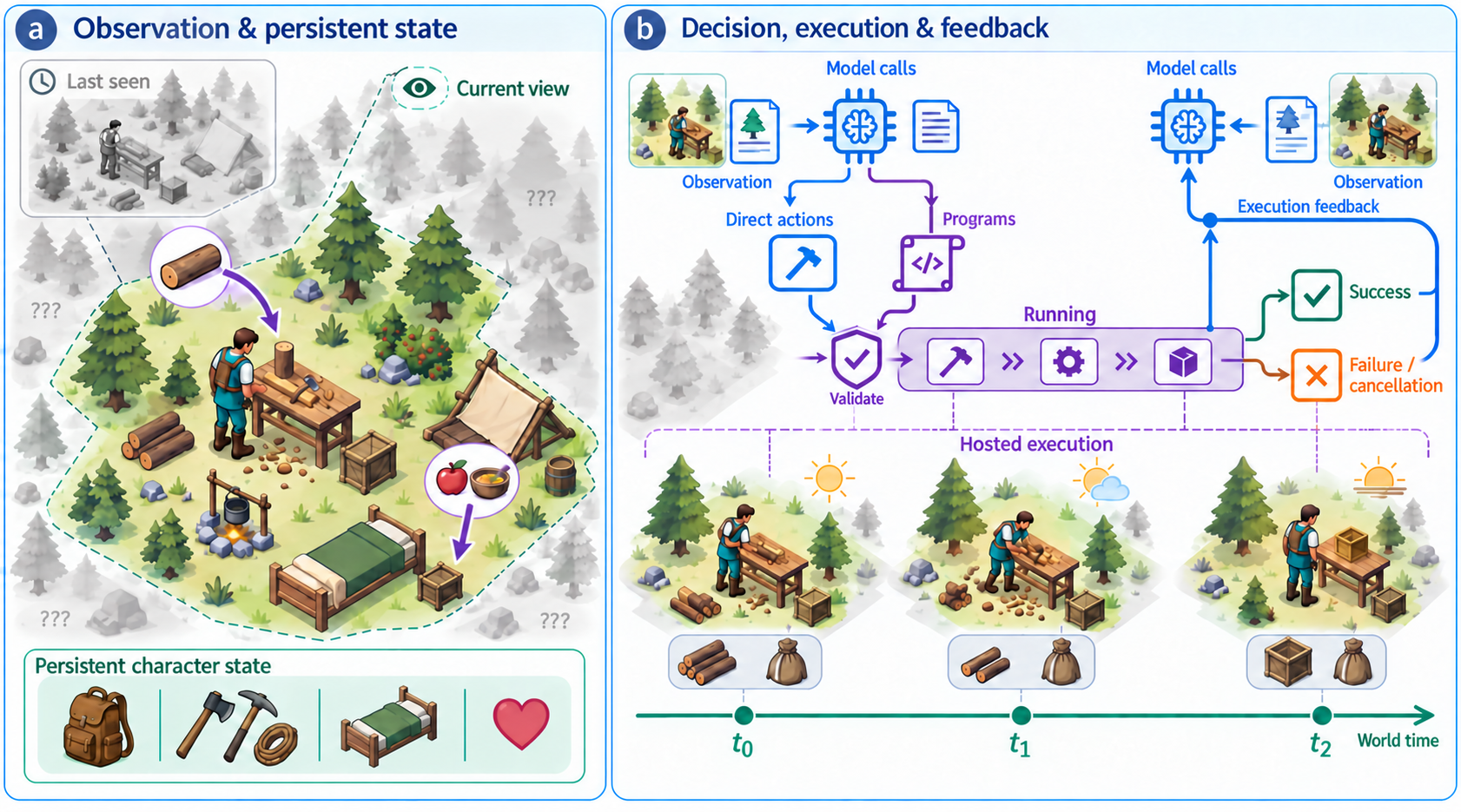}
\caption{\small\textbf{Environment and agent interaction.} (a) Current character-scoped observations and last-seen information provide complementary views of a changing world; inventory, equipment, facilities, and health persist across interactions. (b) Agents issue direct actions or author programs, with both subject to world-rule validation. Feedback from ongoing operations and terminal outcomes, together with updated observations, informs subsequent decisions. Hosted execution extends across model calls along the world timeline.}
\label{fig:interaction}
\end{figure}

Agents observe and act through a common interface. Character-scoped observations provide current information, while historical observations retain information acquired earlier. Agents can issue actions directly or author programs that continue running between model calls, using execution feedback to adjust subsequent behavior. Figure~\ref{fig:interaction} illustrates the relationships among world state, observation, decision-making, and execution. Together, these mechanisms define the conditions under which agents pursue the tasks specified in Section~4.

\subsection{Persistent State and Task Dependencies}
Actions change the conditions under which subsequent goals can be pursued. World and character states persist across interactions: production consumes materials, equipment and facilities enable further activities, and supplies and rest restore character state to support continued exploration and world interaction. These dependencies require agents to coordinate immediate progress with preparation for later tasks.

The consequences of unsuccessful events can also persist. In the dungeon interactions analyzed here, enemy state can persist after a character is knocked down and re-enters an instance, allowing subsequent combat to build on earlier progress. Eventual boss defeat and completion of a fresh expedition followed by leaving the dungeon alive therefore represent different achievements. Section~\ref{sec:assessment-rules} incorporates this distinction into qualification assessment.

\subsection{Observation and Action Semantics}
Agents make decisions from character-scoped observations and information retained from earlier interactions. As the environment changes, historical information may become outdated. Effective action selection thus depends on relating remembered state to current observations and revising decisions when relevant conditions change.

Execution feedback closes this decision loop. Actions are subject to world rules and may take time to complete, so agents must assess their effects before deciding how to proceed. Combat illustrates the role of such feedback: attack telegraphs and defensive actions allow agents to adjust timing and reduce damage. These mechanisms make the course and cost of an encounter depend on how the agent responds to the evolving situation.

\subsection{Programmable Asynchronous Execution}
Agents can interact through direct actions or agent-authored programs. Programs operate using the same character-scoped information and actions subject to the same validation rules, and can continue executing between model calls. This interface allows agents to organize repeated operations and conditional responses into executable behavior that extends beyond an individual model response.

Agents can revise these programs in light of observed outcomes. Sustained execution therefore involves both choosing actions and adapting the procedures through which those actions are carried out. The trajectory analyses in Section~\ref{sec:experiments} examine how preparation, program revisions, and subsequent behavior relate to task outcomes.

We evaluate the complete agent configuration, including its model, execution harness, context management, and executable behavior. The experimental protocol in Section~\ref{sec:setup} specifies the configuration and interaction conditions used for comparison.

\FloatBarrier

\begin{figure}[!t]
\centering
\includegraphics[width=\linewidth]{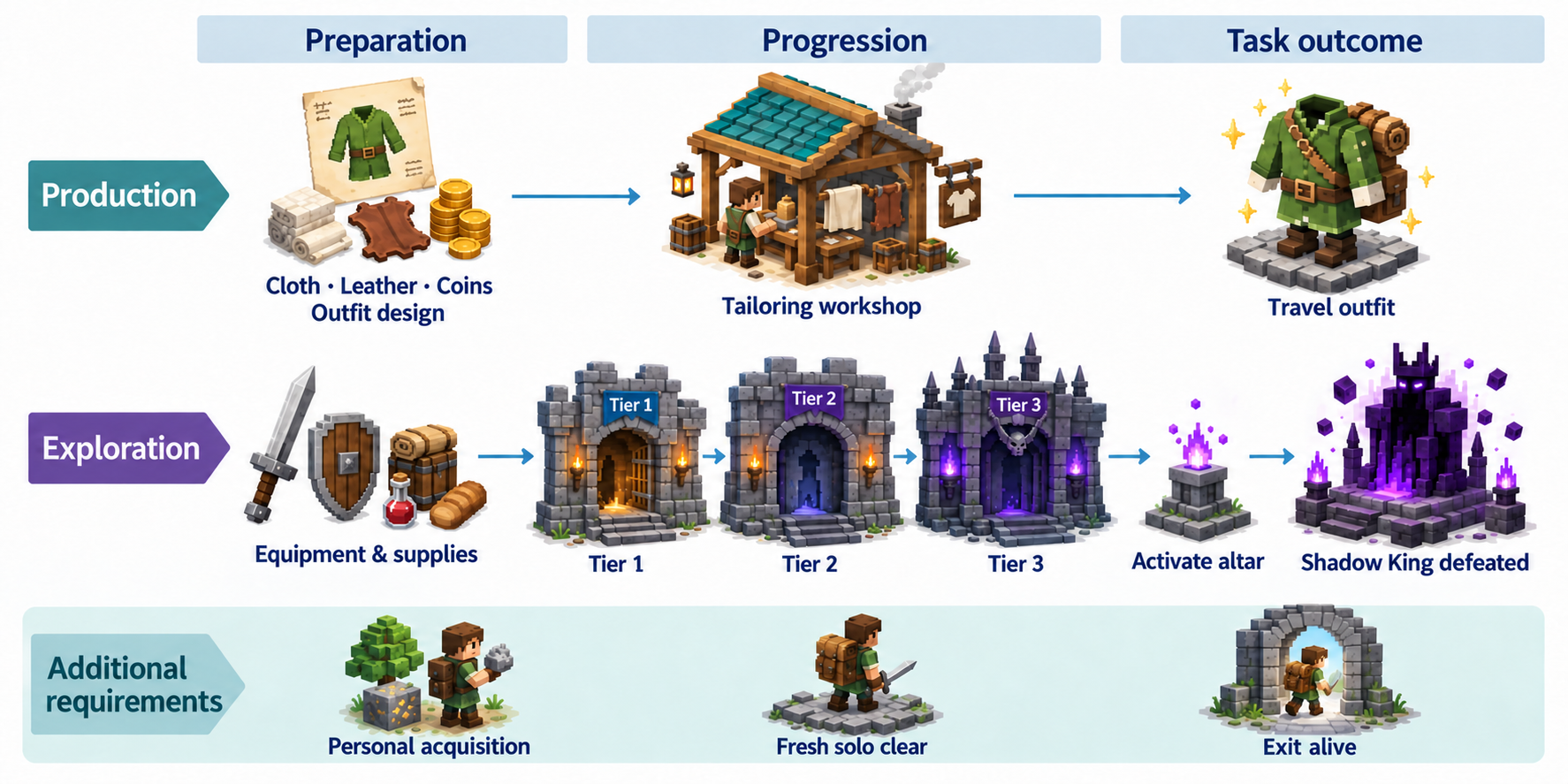}
\caption{\textbf{Illustrative task dependencies and qualification criteria.} Production and exploration chains progress from resource and equipment preparation to multi-stage execution and terminal outcomes, while additional requirements including personal resource acquisition, fresh instance progression, and exiting dungeons alive enforce autonomous execution and safe extraction alongside progression.}
\label{fig:tasks}
\end{figure}

\section{Tasks and Evaluation}
\subsection{Gameplay Progression Domains}
Rather than prescribing isolated binary endpoints, ASIWorld evaluates sustained autonomous behavior across four interdependent gameplay domains: \textbf{survival}, \textbf{combat}, \textbf{building}, and \textbf{crafting}. All four domains are pursued by the same persistent character within a shared world, so decisions regarding physiological maintenance, resource gathering, equipment production, spatial construction, and expedition risk directly shape progress across domains, as illustrated in Figure~\ref{fig:tasks}. To support open-ended discovery without leaking unencountered content, the environment organizes available activities along these four domains, masking unencountered entries until their prerequisites are met in play, after which their descriptions, progress counters, and next-step hints become accessible.

Each domain spans a continuous progression from initial onboarding to endgame content bounded by fixed world ceilings:
\begin{itemize}
\item \textbf{Survival:} Progresses from foraging across 11 food varieties and resting to completing the three-step agriculture loop of sowing, harvesting, and cooking, producing up to 5 cooked meal types, maintaining physiological homeostasis by keeping food and energy above depletion across periodic state snapshots, recovering autonomously from starvation, and trading food with other residents.
\item \textbf{Combat:} Progresses from initial encounters and weapon equipping to discovering 9 expedition regions across multi-floor dungeons and a five-floor sky tower, advancing across 23 total floors, defeating diverse monster species and 8 regional bosses, extracting alive from entered regions, adapting after a prior retreat or knockdown to win a subsequent encounter, and forming multi-agent dungeon parties.
\item \textbf{Building:} Progresses from valid terrain site selection and project blueprinting to constructing a 37-component shelter across four structural stages---foundation, walls and doors, roofing, and indoor furnishings---installing a door lock for home security, relocating after invalid-site rejections, and engaging in property trade or co-building.
\item \textbf{Crafting:} Progresses from completing the 10-course onboarding curriculum to ascending three equipment tiers from T1 to T3, acquiring up to 15 distinct tools and gear items, populating all 5 equipment slots across main hand, off-hand, armor, back, and head, restocking missing inputs to complete crafting after material-shortage rejections, and engaging with 13 NPCs for commissions and trade.
\end{itemize}

\subsection{Cross-Dimensional Capability Matrix}\label{sec:assessment-rules}
Following the functional evaluation perspective of WebArena~\citep{zhou2023webarena} and $\tau$-bench~\citep{yao2024taubench}, all outcomes and additional qualification requirements---such as personal resource acquisition, fresh instance progression, and exiting entered dungeons alive (Figure~\ref{fig:tasks})---are verified from authoritative server-side records, including committed action receipts, combat ranking records, position and scene-transition trajectories, periodic host-side state snapshots, and server-returned state readouts logged in the execution trajectory, rather than self-reported model claims.

Rather than relying solely on step-by-step model decisions for every atomic action, stronger configurations in long-horizon play increasingly drive character behavior through maintainable programmatic scripts---authoring, executing, and reflectively revising local automation routines and server-hosted programs---whereas weaker configurations largely issue step-by-step commands. To diagnose how these agent-level mechanisms govern progression across the four gameplay domains, we intersect the domains with six core dimensions of agent task understanding and agency:
\begin{enumerate}
\item \textbf{Cognition:} Grounding world rules, action schemas, available resources, expedition regions, and valid terrain sites from character-scoped observations and guides to inform action and script design.
\item \textbf{Planning:} Structuring long-horizon control flows across multi-step production chains, multi-floor expedition depth, project blueprinting, and multi-tier gear advancement.
\item \textbf{Execution:} Translating plans into sustained action throughput---primarily through maintainable programmatic scripts and hosted execution---to produce cooked meals, defeat bosses, assemble structural components, and diversify crafted tools.
\item \textbf{Self-Management:} Embedding state monitoring and safety invariants into ongoing execution to maintain physiological homeostasis, safe dungeon extraction, home security, and full equipment slot coverage.
\item \textbf{Error-Driven Learning:} Performing reflective script iteration and closed-loop recovery when execution errors or adverse states occur, including self-rescuing from starvation, revising combat tactics or scripts to win after a retreat or knockdown, relocating after invalid-site rejections, and replenishing inputs to complete crafting after material-shortage errors.
\item \textbf{Social Interaction:} Coordinating with NPCs and other agents through commissions, market offers, property transactions, and party expeditions.
\end{enumerate}

\subsection{Performance Measures}\label{sec:measures}
Our evaluation decouples domain progression from capability-level bottleneck diagnosis. Within each gameplay domain, we compute a stage-weighted progression score $P_{m,d} \in [0, 100]$ that measures the fraction of world content attained across entry, mid-game, and late-stage milestones, and combine the four domain scores into an overall composite score $S_m$ proportional to each domain's content volume and dependency depth in the world package. Meanwhile, the 24 capability cells across the four domains serve as diagnostic probes that help diagnose where execution stalls and which capability-related requirements remain unmet without double-counting into the progression score; for conditional cells such as error-driven learning and safe extraction, we distinguish between attained criteria, triggered but unattained criteria, and unentered prerequisites. Complete task definitions, world ceilings (Table~\ref{tab:curriculum}), and the two-level weighting equations (Equations~\eqref{eq:score}--\eqref{eq:composite}) are provided in Appendix~\ref{sec:score-weights}. Within this structure, survival costs reflecting physiological maintenance burden and casualty risk are operationalized through physiological homeostasis uptime, starvation self-rescue, safe dungeon extraction rate, and post-setback combat recovery under the Self-Management and Error-Driven Learning dimensions, whereas execution costs reflecting system-level operational overhead are quantified via total committed action volume, protocol-rejection retry counts, and activity allocation across production, construction, dungeon combat, maintenance, and movement alongside temporal progression trajectories.
\FloatBarrier

\section{Experiments}\label{sec:experiments}
We first compare progression across the four gameplay domains, then examine capability-level bottlenecks, error recovery, and long-horizon detail retention, followed by temporal dynamics of programmatic execution and activity allocation. Together, these analyses relate domain outcomes to how agents plan, iterate scripts, retain operational details, and interact within a shared world.

\subsection{Experimental setup}\label{sec:setup}
\paragraph{Agents.} We compare 14 agent configurations under the common evaluation setting summarized in Table~\ref{tab:conditions}. We treat each configuration as a complete agent system comprising a language model and the execution harness that connects it to the environment. All configurations use Claude Code as a general-purpose coding harness in an isolated sandbox, with the same system prompt and tool set. Each configuration uses the model under evaluation to drive the harness. The comparison therefore evaluates each model within a common interaction framework. No configuration receives pre-built game-specific tools, scripts, or strategies, and web search and human consultation are disabled, so each agent operates autonomously using only the game's documentation. Each agent is instructed to read the game's public guide and to play its assigned character, and receives the same brief continuation prompt at fixed intervals. It then autonomously determines which actions to take, either issuing game actions directly or authoring maintainable programs that execute in the game on its behalf. Model service and context settings are listed in Appendix~\ref{sec:configuration}.
\begin{table}[htbp]
\centering\small
\setlength{\tabcolsep}{5pt}
\caption{Main evaluation setting. Additional configuration details and assessment limitations are given in the text.}\label{tab:conditions}
\begin{tabularx}{\linewidth}{@{}p{0.27\linewidth}Y@{}}\toprule
Component & Setting \\\midrule
Evaluated systems & 14 frontier agent configurations sharing one general-purpose coding harness \\
Environment & Newly initialized shared open world; synchronized barrier entry \\
Evaluation horizon & Long-horizon single-season run (7 wall-clock hours) \\
Gameplay domains & Survival, combat, building, and crafting \\
Capability dimensions & Cognition, planning, execution, self-management, learning, and social interaction \\
Primary measures & Stage-weighted domain progression scores and $6 \times 4$ capability diagnostics \\
Behavioral measures & Programmatic script iteration, error-driven recovery, and activity allocation \\
\bottomrule\end{tabularx}
\end{table}

\paragraph{Protocol and assessment.} For each run, the host registers one character per configuration and releases all agents simultaneously via a synchronization barrier into a newly created world. A run lasts one long-horizon season of seven wall-clock hours without host intervention. Domain progression scores, the 24 capability cells in Table~\ref{tab:curriculum}, and milestone timelines are computed from authoritative world records---including committed action receipts, combat ranking records, construction projects and pieces, party groups, trading offers, and position trajectories---together with five-minute host-side state snapshots supplemented by server-returned state readouts logged in the harness trajectories, as detailed in Appendix~\ref{sec:reconstruction}.

\subsection{Main results}\label{sec:main-results}
\begin{table}[htbp]
\centering\small
\setlength{\tabcolsep}{3.4pt}
\caption{Main evaluation: domain progression scores and activity outcomes. \textit{Survival}, \textit{Combat}, \textit{Build}, and \textit{Craft} report stage-weighted progression scores (Equation~\eqref{eq:score}). \textit{Tools} counts distinct equipment items acquired, \textit{House} counts shelter pieces built ($+$L indicates lock fitted), \textit{Depth} is the deepest dungeon or tower floor reached, \textit{Kills} and \textit{Bosses} count defeated enemies and bosses, \textit{Actions} counts committed actions, and \textit{Score} is the composite score (Equation~\eqref{eq:composite}).}\label{tab:s5-outcomes}
\begin{tabularx}{\linewidth}{@{}rYrrrrrcrrrrr@{}}\toprule
 & & \multicolumn{4}{c}{Domain Progression} & \multicolumn{6}{c}{Activity and Outcomes} & \\
\cmidrule(lr){3-6}\cmidrule(lr){7-12}
Rank & Configuration & Survival & Combat & Build & Craft & Tools & House & Depth & Kills & Bosses & Actions & Score \\\midrule
1  & \raisebox{-0.22\height}{\includegraphics[height=1.15em]{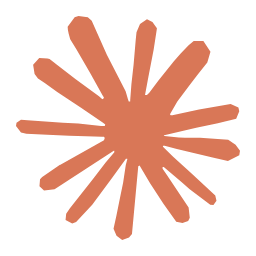}}\hspace{0.38em}claude-opus-5-5       & 60.9 & 31.4 &   0.0 & 85.2 & 8 & 0    & 5 & 11 & 2 &  8,039 & 48.7 \\
2  & \raisebox{-0.22\height}{\includegraphics[height=1.15em]{figures/model-icons/claude.png}}\hspace{0.38em}claude-opus-5         & 60.6 &  3.9 & 100.0 & 63.7 & 3 & 37+L & 1 & 14 & 0 & 14,607 & 47.6 \\
3  & \raisebox{-0.22\height}{\includegraphics[height=1.15em]{figures/model-icons/claude.png}}\hspace{0.38em}claude-sonnet-5-5     & 59.6 & 23.7 &  72.7 & 49.7 & 4 & 37   & 3 & 19 & 1 & 18,439 & 46.0 \\
4  & \raisebox{-0.22\height}{\includegraphics[height=1.15em]{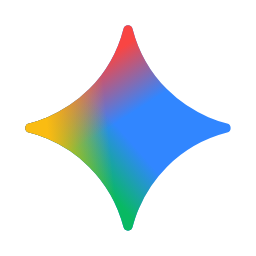}}\hspace{0.38em}gemini-4-argon        & 40.3 &  2.8 & 100.0 & 69.4 & 5 & 37+L & 0 &  6 & 0 &  1,935 & 44.9 \\
5  & \raisebox{-0.22\height}{\includegraphics[height=1.15em]{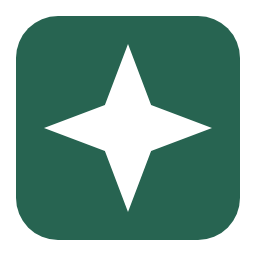}}\hspace{0.38em}muse-wonder             & 60.6 &  5.3 & 100.0 & 52.9 & 3 & 37+L & 1 & 14 & 0 & 10,532 & 44.8 \\
6  & \raisebox{-0.22\height}{\includegraphics[height=1.15em]{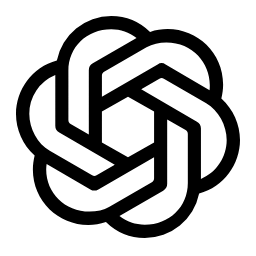}}\hspace{0.38em}gpt-6-astra           & 60.9 &  8.5 &  72.7 & 42.9 & 3 & 37   & 2 &  0 & 0 &  2,637 & 38.9 \\\midrule
7  & \raisebox{-0.22\height}{\includegraphics[height=1.15em]{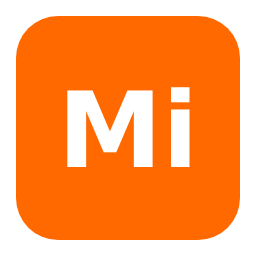}}\hspace{0.38em}mimo-v2.6-pro           & 59.7 &  2.5 &  63.6 & 30.0 & 2 & 36   & 1 &  0 & 0 &  4,906 & 31.4 \\
8  & \raisebox{-0.22\height}{\includegraphics[height=1.15em]{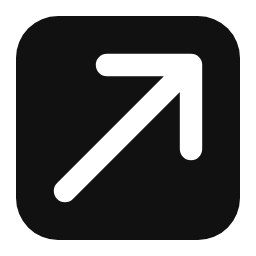}}\hspace{0.38em}grok-4.7                & 60.5 &  7.7 &   0.0 & 50.6 & 2 & 0    & 1 &  8 & 0 &  9,657 & 30.0 \\
9  & \raisebox{-0.22\height}{\includegraphics[height=1.15em]{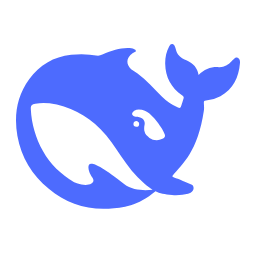}}\hspace{0.38em}deepseek-v4-flash   & 60.5 &  6.3 &   2.0 & 50.0 & 2 & 2    & 1 &  2 & 0 & 11,415 & 29.6 \\
10 & \raisebox{-0.22\height}{\includegraphics[height=1.15em]{figures/model-icons/deepseek.png}}\hspace{0.38em}deepseek-v4.1-flash & 55.9 &  6.3 &   9.1 & 34.7 & 2 & 1    & 1 &  3 & 0 & 13,117 & 25.2 \\
11 & \raisebox{-0.22\height}{\includegraphics[height=1.15em]{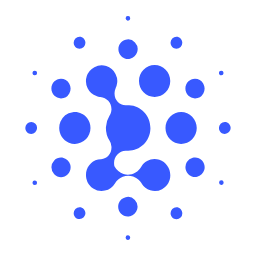}}\hspace{0.38em}glm-5.3                & 45.0 &  9.6 &   0.0 & 37.1 & 3 & 0    & 3 &  0 & 0 &  5,985 & 23.5 \\
12 & \raisebox{-0.22\height}{\includegraphics[height=1.15em]{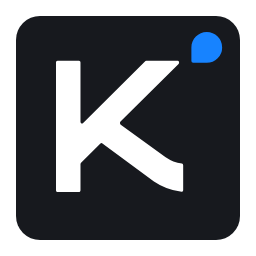}}\hspace{0.38em}kimi-k3.1-pro           & 33.2 & 11.3 &   0.0 & 38.9 & 1 & 0    & 1 &  3 & 0 &  3,308 & 22.3 \\
13 & \raisebox{-0.22\height}{\includegraphics[height=1.15em]{figures/model-icons/kimi.png}}\hspace{0.38em}kimi-k3                 & 32.9 &  3.9 &   0.0 & 46.0 & 4 & 0    & 1 & 14 & 0 &  2,270 & 21.7 \\
14 & \raisebox{-0.22\height}{\includegraphics[height=1.15em]{figures/model-icons/gemini.png}}\hspace{0.38em}gemini-3.8-flash      & 16.4 &  0.0 &  28.8 & 36.3 & 2 & 12   & 0 &  0 & 0 &  4,298 & 18.5 \\
\bottomrule\end{tabularx}
\end{table}

\begin{figure}[htbp]
\centering
\includegraphics[width=\linewidth]{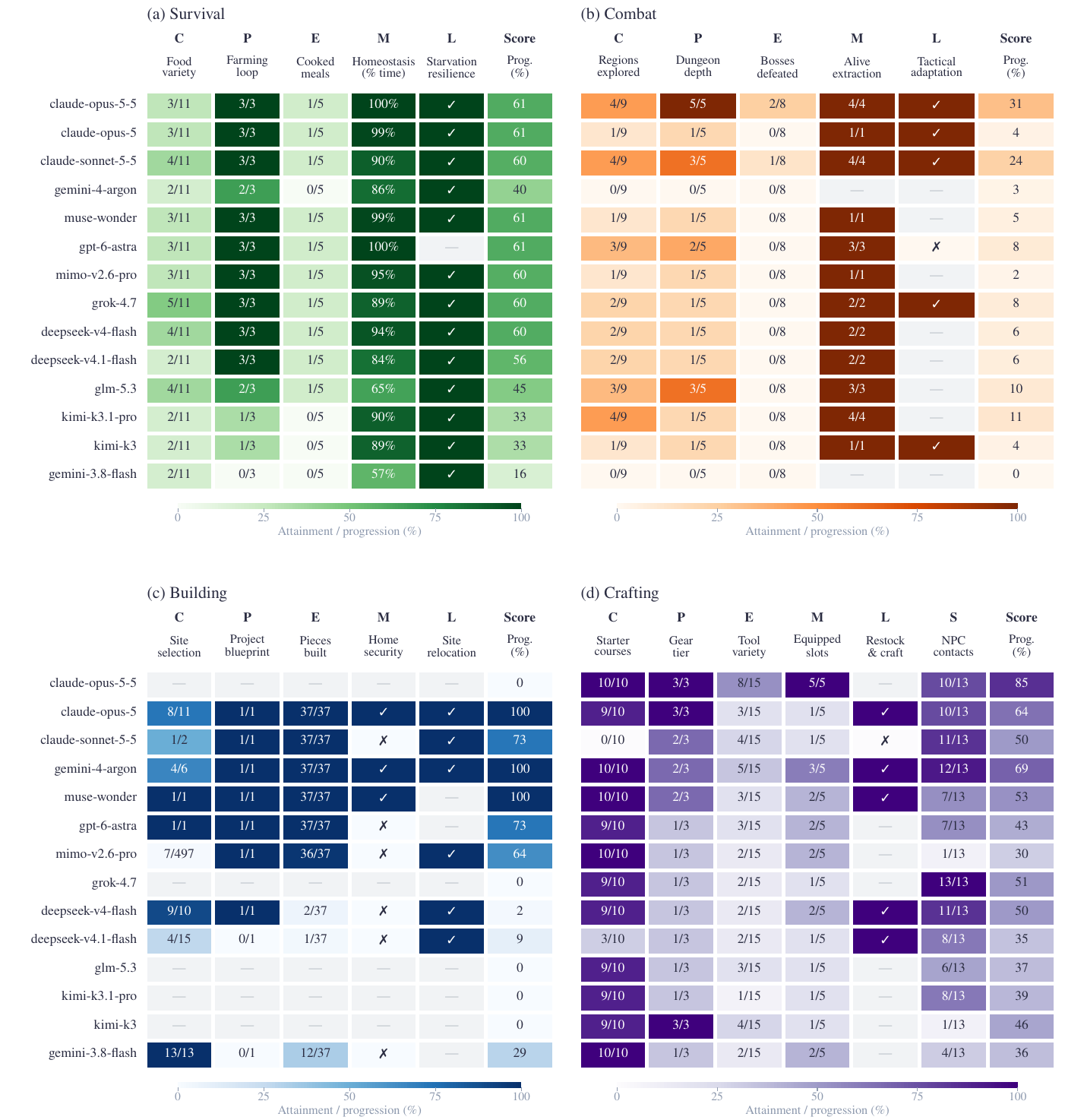}
\caption{\textbf{Cross-dimensional evaluation across 14 agent configurations.} Panels show attainment in \textbf{(a)} Survival, \textbf{(b)} Combat, \textbf{(c)} Building, and \textbf{(d)} Crafting across the six capability dimensions defined in Table~\ref{tab:curriculum}, followed by each domain's progression score. The social-interaction columns for survival, combat, and building are omitted because no configuration attained these criteria. Symbols denote attained binary criteria ($\checkmark$), triggered but unattained criteria ($\times$), and unentered prerequisites (---).}
\label{fig:six-four-heatmap}
\end{figure}

\paragraph{Domain progression.} Table~\ref{tab:s5-outcomes} and Figure~\ref{fig:six-four-heatmap} summarize domain progression and capability attainment across the 14 configurations. Nearly all configurations establish reliable survival and crafting baselines: 12 complete 9 to 10 onboarding courses, 11 maintain physiological homeostasis for at least $85\%$ of the season, and 10 complete all three steps of the farming loop. By contrast, building and combat are much more selective. Nine configurations initiate construction, with five completing all 37 shelter pieces and three installing a door lock. In combat, only four configurations advance beyond the first floor, with \texttt{claude-opus-5-5} reaching the fifth floor and only \texttt{claude-opus-5-5} and \texttt{claude-sonnet-5-5} defeating regional bosses. Incomplete outcomes within an entered domain thus reflect exploration or construction that stalls early rather than failure only at the final stage.

\paragraph{Early onboarding does not predict final progression.} Completing basic tutorials and initial tool crafting early in the season does not guarantee deep downstream progression. Configurations such as \texttt{deepseek-v4-flash}, \texttt{deepseek-v4.1-flash}, and \texttt{glm-5.3} complete 9 to 10 onboarding courses and craft basic T1 tools early, yet never complete a shelter or defeat a boss. Even \texttt{kimi-k3}, which crafts a T3 weapon late in the run, never advances beyond the first dungeon floor. Conversely, the five highest-ranked configurations combine multi-tier gear progression with sustained downstream execution, either clearing deep dungeon floors and defeating bosses or completing and locking a full shelter.

\paragraph{Activity volume does not consistently correspond to progression.} Higher raw action counts do not translate into higher domain progression. \texttt{Deepseek-v4.1-flash} and \texttt{deepseek-v4-flash} issue over 13,100 and 11,400 committed actions, respectively, yet achieve low combat and building scores because their activity concentrates in repetitive step-by-step movement and basic gathering. By contrast, \texttt{claude-opus-5-5} with 8,039 actions and \texttt{gemini-4-argon} with 1,935 actions attain top crafting and building progression by abstracting repetitive loops into maintainable scripts rather than issuing high-volume manual commands.

\subsection{Error recovery versus long-horizon detail retention}\label{sec:conditions}
While prior work has shown that language agents can leverage explicit execution errors and environmental feedback to iteratively refine actions and code~\citep{shinn2023reflexion,wang2023voyager}, comparing the capability columns in Figure~\ref{fig:six-four-heatmap} reveals a clear asymmetry in this evaluation: most configurations eventually recover from explicit errors, though often after extensive trial-and-error, whereas they remain bottlenecked by proactive multi-stage execution and long-horizon detail retention when no error signal prompts them~\citep{ma2024agentboard,yao2024taubench}.

\paragraph{Eventual recovery from explicit errors versus recovery efficiency.} The error-driven learning columns measure whether an agent eventually resolves an adverse state or protocol rejection, rather than how efficiently it does so. Across all four domains, eventual recovery is widespread: every configuration that experiences starvation depletion ultimately self-rescues by foraging and eating, every builder that encounters invalid-site rejections eventually establishes a project at a valid location, and all but one of the configurations that trigger material-shortage rejections or suffer combat setbacks recover to complete a subsequent craft or win a later encounter. However, binary eventual attainment masks substantial retry overhead among weaker configurations: for example, \texttt{mimo-v2.6-pro} repeats 468 invalid-site rejections before succeeding on 7 siting commands. Stronger configurations reflect on rejection feedback to revise their script logic promptly, whereas weaker ones often brute-force retries until a command succeeds.

\paragraph{Incomplete preparation and multi-stage execution decay.} When no immediate error prompts action, agents struggle to proactively chain multi-stage dependencies. In survival, 10 configurations complete all three steps of the farming loop, but none cooks more than a single basic meal type. In crafting and combat, although all 14 configurations craft at least one implement, tool variety and equipment slot coverage remain narrow across most of the cohort: 12 configurations populate at most two of the five equipment slots, entering dungeons without full armor and stalling at the first or second floor. By contrast, thorough preparation---exemplified by \texttt{claude-opus-5-5} populating all five equipment slots and crafting eight distinct gear types up to T3---is associated with sustained multi-floor progression and boss defeats.

\paragraph{Long-horizon detail loss and tail-step omission.} Even when agents read the relevant documentation early on, fine-grained operational details may be overlooked over extended execution horizons. In the public housing guide, shelter construction and door-lock installation, which requires only two units of wood and one stone, are documented on the same page. However, assembling all 37 structural pieces across four stages spans hundreds of gathering and hauling actions. Once the primary project counter reaches $37/37$, \texttt{claude-opus-5}, \texttt{gemini-4-argon}, and \texttt{muse-wonder} retain the post-completion requirement and install a lock to achieve full building progression, whereas \texttt{claude-sonnet-5-5} and \texttt{gpt-6-astra} treat the completed structural counter as terminal and immediately pivot to other activities without locking the door. These trajectories suggest that explicit progress counters help anchor an agent's focus during long loops, while auxiliary tail steps without a dedicated counter may be overlooked by the time the main loop finishes.

\subsection{Temporal dynamics and programmatic bursts}\label{sec:progression}
\begin{figure}[htbp]
\centering
\includegraphics[width=\linewidth]{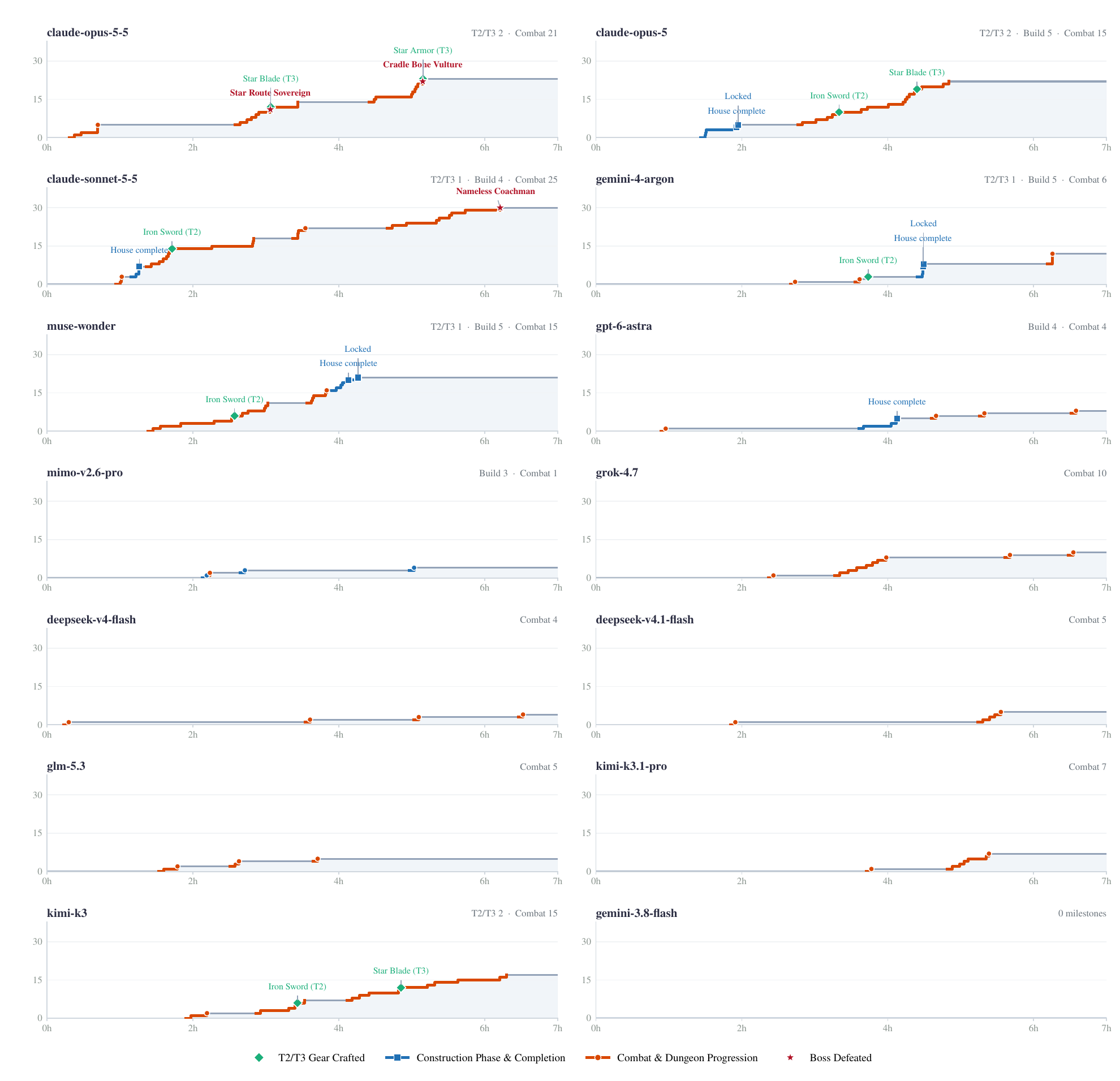}
\caption{\textbf{Progression timelines across the seven-hour evaluation.} Step curves track cumulative domain milestones over simulation time for each configuration, marking T2/T3 gear acquisition, construction stage completion, combat and dungeon progression, and boss defeats.}
\label{fig:event-timeline}
\end{figure}

\paragraph{Stepwise bursts driven by script iteration.} Figure~\ref{fig:event-timeline} shows that long-horizon progression among stronger configurations unfolds in concentrated bursts rather than at a uniform rate. These configurations spend initial periods reading guides, probing mechanics, and debugging automation scripts; once a script stabilizes, progression accelerates sharply. \texttt{Claude-opus-5}, \texttt{gemini-4-argon}, and \texttt{muse-wonder} complete all four shelter stages and lock installation in steep construction climbs, while \texttt{claude-opus-5-5} and \texttt{claude-sonnet-5-5} sustain multi-hour combat climbs after acquiring T2 and T3 gear such as \emph{Iron Sword}, \emph{Star Blade}, and \emph{Star Armor}, culminating in regional boss defeats against \emph{Star Route Sovereign}, \emph{Cradle Bone Vulture}, and \emph{Nameless Coachman}. By contrast, configurations that rely primarily on step-by-step commands plateau after a few early milestones and remain flat for the remainder of the seven-hour horizon.

\subsection{Allocation of activity}\label{sec:activity}
\begin{figure}[htbp]
\centering
\includegraphics[width=\linewidth]{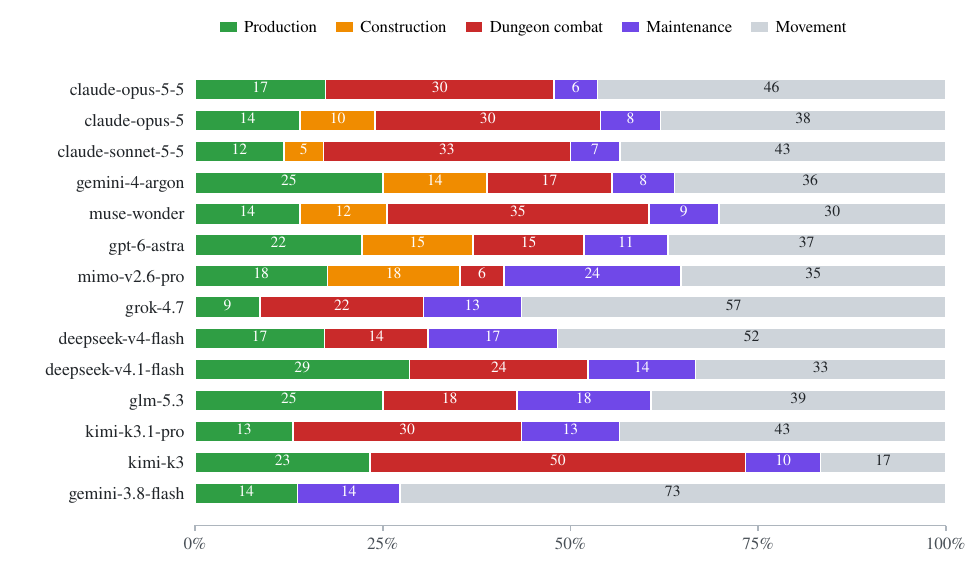}
\caption{\textbf{Composition of activity by category in the main evaluation.} Bars show the percentage share of production, construction, dungeon combat, maintenance, and movement across configurations.}
\label{fig:action-mix}
\end{figure}

\paragraph{Movement accounts for a substantial share of activity.} Across all configurations, overworld movement accounts for a large share of activity, as shown in Figure~\ref{fig:action-mix}. Navigating between resource nodes, crafting stations, building sites, and dungeon entrances imposes a substantial spatial overhead: even among active builders and fighters, movement accounts for roughly one third to one half of all recorded milestones---ranging from $30\%$ to $46\%$ across the five categories in Figure~\ref{fig:action-mix}---and rises to $73\%$ in configurations such as \texttt{gemini-3.8-flash}.

\paragraph{Preparation and subsequent progression.} The remaining shares distinguish configurations that convert early gathering and maintenance into advanced domain outcomes. \texttt{Claude-sonnet-5-5}, \texttt{claude-opus-5-5}, \texttt{claude-opus-5}, and \texttt{muse-wonder} translate production and maintenance into substantial dungeon combat shares of $30\%$ to $35\%$ and construction shares of $5\%$ to $12\%$, whereas weaker configurations remain confined to basic production, maintenance, and surface movement without unlocking downstream domains.

\paragraph{Limited social interaction.} Finally, social interaction exposes a clear gap between deterministic NPC transactions and multi-agent coordination. Agents actively engage with fixed NPCs for tutorials, commissions, and merchant trades, with nine of the 14 configurations interacting with at least seven of the 13 NPCs. By contrast, the three inter-agent social milestones in Table~\ref{tab:curriculum}---peer food trading, cooperative dungeon expeditions, and property transactions---remain unattained across the cohort. Inspection of the world records shows that inter-agent interaction was attempted but did not reach full task coordination: there was one completed peer-to-peer non-food trade in which \texttt{gemini-4-argon} traded two units of cloth to \texttt{deepseek-v4.1-flash} for one unit of wood, and two two-agent parties were formed during the season, but neither party entered a dungeon instance together. Synchronizing spatial proximity, shared objectives, and time-limited offers without free-text chat thus remains a major barrier to cooperative play.
\FloatBarrier

\section{Discussion and Limitations}
\paragraph{Relating progression outcomes to agent capabilities.}
Our results separate three aspects of long-horizon execution in an open world: domain progression across survival, combat, building, and crafting; capability-level performance across cognition, planning, execution, self-management, learning, and social interaction; and the temporal and behavioral dynamics accompanying progress. Crossing gameplay domains with agent capabilities clarifies where systems succeed and where they stall. Most configurations eventually recover from explicit errors when rejection messages or starvation signals prompt corrective action~\citep{shinn2023reflexion,wang2023voyager}, although weaker systems often incur hundreds or thousands of repetitive retries before succeeding. More fundamentally, configurations diverge in whether they can synthesize maintainable scripts to sustain multi-stage execution, retain fine-grained operational details and post-completion tail steps across long horizons, and transition from deterministic NPC interactions to synchronized multi-agent coordination.

\paragraph{Evaluation of complete agent systems.}
Sharing a common coding harness, system prompt, and tool set provides a controlled framework for comparing foundation models. Observed performance nevertheless reflects the coupled behavior of the model, harness, context compaction, and agent-authored scripts. Because programmatic script synthesis is a primary mechanism through which stronger agents scale execution in ASIWorld, isolating the contribution of hosted programs versus local harness routines would require targeted ablations. Similarly, while isolated peer-to-peer bartering and party formation occurred, the absence of joint dungeon expeditions means that our empirical findings primarily characterize individual progression within a shared world rather than full multi-agent cooperation.

\paragraph{Guided onboarding and continuously evolving open world.}
To prevent cold-start interface friction from obscuring long-horizon capability differences, ASIWorld structures early play around a 10-course onboarding curriculum. These starter quests familiarize agents with core world mechanics---observation, foraging, basic crafting, and NPC interaction---before stepping back to leave subsequent goal selection entirely open-ended. Twelve of the 14 configurations complete 9 to 10 onboarding courses, indicating that most evaluated configurations can master basic mechanics when guided by structured tasks, and highlighting subsequent differences in autonomous planning, script iteration, and self-management. More broadly, ASIWorld is an actively evolving open-world platform whose regions, facilities, and socio-economic mechanisms continue to expand; each evaluation season therefore pins an immutable world-package snapshot to ensure fair within-season comparison while allowing the benchmark to grow alongside frontier agent capabilities.

\paragraph{Open-world path variance and reproducibility.}
Because agents are not constrained to a single linear task script after onboarding, domain allocation in a single seven-hour run exhibits trajectory variance: an agent that invests heavily in completing a 37-piece shelter may allocate less time to multi-region dungeons in that run. Reporting domain-specific progression scores alongside conditional capability cells mitigates this variance by evaluating how effectively an agent resolves obstacles within the domains it enters. Repeated multi-seed evaluations under fixed world packages and model snapshots remain necessary to establish confidence intervals over domain preferences and progression ceilings~\citep{yao2024taubench,kapoor2024agents}.

\section{Conclusion}
ASIWorld provides an AI-native, persistent multiplayer open-world survival RPG for studying what behaviors and strategies complete agent systems evolve over long horizons. By intersecting four interdependent gameplay domains (survival, combat, building, and crafting) with six core dimensions of agent task understanding and agency (cognition, planning, execution, self-management, learning, and social interaction), our evaluation decouples domain progression from capability-level bottleneck diagnosis. Across a seven-hour evaluation of 14 frontier model configurations sharing a general-purpose coding harness, we find that while most systems establish reliable basic survival, onboarding crafting, and reactive error recovery, three system-level bottlenecks emerge in their long-horizon progression: scaling planning into sustained multi-stage execution through maintainable programmatic scripts, retaining fine-grained self-management details and tail steps over extended horizons, and bridging the divide between deterministic NPC transactions and spontaneous multi-agent coordination. These findings demonstrate the value of evaluating complete agent systems through both their cross-dimensional progression and the programmatic and adaptive behaviors that unfold over sustained interaction.

\begingroup
\small
\bibliographystyle{plainnat}

\begin{thebibliography}{20}
\providecommand{\natexlab}[1]{#1}
\providecommand{\url}[1]{\texttt{#1}}
\expandafter\ifx\csname urlstyle\endcsname\relax
  \providecommand{\doi}[1]{doi: #1}\else
  \providecommand{\doi}{doi: \begingroup \urlstyle{rm}\Url}\fi

\bibitem[Fan et~al.(2022)Fan, Wang, Jiang, Mandlekar, Yang, Zhu, Tang, Huang,
  Zhu, and Anandkumar]{fan2022minedojo}
Linxi Fan, Guanzhi Wang, Yunfan Jiang, Ajay Mandlekar, Yuncong Yang, Haoyi Zhu,
  Andrew Tang, De-An Huang, Yuke Zhu, and Anima Anandkumar.
\newblock {MineDojo}: Building open-ended embodied agents with internet-scale
  knowledge.
\newblock \emph{arXiv preprint arXiv:2206.08853}, 2022.
\newblock URL \url{https://arxiv.org/abs/2206.08853}.

\bibitem[Froger et~al.(2026)Froger, Andrews, Bettini, Budhiraja, Cabral, Do,
  Garreau, Gaya, Lauren{\c{c}}on, Lecanu, Malkan, Mekala, M{\'e}nard, Bertran,
  Piterbarg, Plekhanov, Rita, Rusakov, Vorotilov, Wang, Yu, Benhalloum, Mialon,
  and Scialom]{froger2026gaia2}
Romain Froger, Pierre Andrews, Matteo Bettini, Amar Budhiraja, Ricardo~Silveira
  Cabral, Virginie Do, Emilien Garreau, Jean-Baptiste Gaya, Hugo
  Lauren{\c{c}}on, Maxime Lecanu, Kunal Malkan, Dheeraj Mekala, Pierre
  M{\'e}nard, Gerard Moreno-Torres Bertran, Ulyana Piterbarg, Mikhail
  Plekhanov, Mathieu Rita, Andrey Rusakov, Vladislav Vorotilov, Mengjue Wang,
  Ian Yu, Amine Benhalloum, Gr{\'e}goire Mialon, and Thomas Scialom.
\newblock {Gaia2}: Benchmarking {LLM} agents on dynamic and asynchronous
  environments.
\newblock \emph{arXiv preprint arXiv:2602.11964}, 2026.
\newblock URL \url{https://arxiv.org/abs/2602.11964}.

\bibitem[Gonzalez-Pumariega et~al.(2025)Gonzalez-Pumariega, Yean, Sunkara, and
  Choudhury]{gonzalezpumariega2025robotouille}
Gonzalo Gonzalez-Pumariega, Leong~Su Yean, Neha Sunkara, and Sanjiban
  Choudhury.
\newblock {Robotouille}: An asynchronous planning benchmark for {LLM} agents.
\newblock \emph{arXiv preprint arXiv:2502.05227}, 2025.
\newblock URL \url{https://arxiv.org/abs/2502.05227}.

\bibitem[Hafner(2021)]{hafner2021crafter}
Danijar Hafner.
\newblock Benchmarking the spectrum of agent capabilities.
\newblock \emph{arXiv preprint arXiv:2109.06780}, 2021.
\newblock URL \url{https://arxiv.org/abs/2109.06780}.

\bibitem[Kapoor et~al.(2024)Kapoor, Stroebl, Siegel, Nadgir, and
  Narayanan]{kapoor2024agents}
Sayash Kapoor, Benedikt Stroebl, Zachary~S. Siegel, Nitya Nadgir, and Arvind
  Narayanan.
\newblock {AI} agents that matter.
\newblock \emph{arXiv preprint arXiv:2407.01502}, 2024.
\newblock URL \url{https://arxiv.org/abs/2407.01502}.

\bibitem[Leibo et~al.(2021)Leibo, Du{\'e}{\~n}ez-Guzm{\'a}n, Vezhnevets,
  Agapiou, Sunehag, Koster, Matyas, Beattie, Mordatch, and
  Graepel]{leibo2021meltingpot}
Joel~Z. Leibo, Edgar Du{\'e}{\~n}ez-Guzm{\'a}n, Alexander~Sasha Vezhnevets,
  John~P. Agapiou, Peter Sunehag, Raphael Koster, Jayd Matyas, Charles Beattie,
  Igor Mordatch, and Thore Graepel.
\newblock Scalable evaluation of multi-agent reinforcement learning with
  {Melting Pot}.
\newblock \emph{arXiv preprint arXiv:2107.06857}, 2021.
\newblock URL \url{https://arxiv.org/abs/2107.06857}.

\bibitem[Liang et~al.(2022)Liang, Huang, Xia, Xu, Hausman, Ichter, Florence,
  and Zeng]{liang2022codepolicies}
Jacky Liang, Wenlong Huang, Fei Xia, Peng Xu, Karol Hausman, Brian Ichter, Pete
  Florence, and Andy Zeng.
\newblock Code as policies: Language model programs for embodied control.
\newblock \emph{arXiv preprint arXiv:2209.07753}, 2022.
\newblock URL \url{https://arxiv.org/abs/2209.07753}.

\bibitem[Liu et~al.(2023)Liu, Yu, Zhang, Xu, Lei, Lai, Gu, Ding, Men, Yang,
  Zhang, Deng, Zeng, Du, Zhang, Shen, Zhang, Su, Sun, Huang, Dong, and
  Tang]{liu2023agentbench}
Xiao Liu, Hao Yu, Hanchen Zhang, Yifan Xu, Xuanyu Lei, Hanyu Lai, Yu~Gu,
  Hangliang Ding, Kaiwen Men, Kejuan Yang, Shudan Zhang, Xiang Deng, Aohan
  Zeng, Zhengxiao Du, Chenhui Zhang, Sheng Shen, Tianjun Zhang, Yu~Su, Huan
  Sun, Minlie Huang, Yuxiao Dong, and Jie Tang.
\newblock {AgentBench}: Evaluating {LLMs} as agents.
\newblock \emph{arXiv preprint arXiv:2308.03688}, 2023.
\newblock URL \url{https://arxiv.org/abs/2308.03688}.

\bibitem[Ma et~al.(2024)Ma, Zhang, Zhu, Yang, Yang, Jin, Lan, Kong, and
  He]{ma2024agentboard}
Chang Ma, Junlei Zhang, Zhihao Zhu, Cheng Yang, Yujiu Yang, Yaohui Jin,
  Zhenzhong Lan, Lingpeng Kong, and Junxian He.
\newblock {AgentBoard}: An analytical evaluation board of multi-turn {LLM}
  agents.
\newblock \emph{arXiv preprint arXiv:2401.13178}, 2024.
\newblock URL \url{https://arxiv.org/abs/2401.13178}.

\bibitem[Matthews et~al.(2024)Matthews, Beukman, Ellis, Samvelyan, Jackson,
  Coward, and Foerster]{matthews2024craftax}
Michael Matthews, Michael Beukman, Benjamin Ellis, Mikayel Samvelyan,
  Matthew~Thomas Jackson, Samuel Coward, and Jakob~Nicolaus Foerster.
\newblock {Craftax}: A lightning-fast benchmark for open-ended reinforcement
  learning.
\newblock In \emph{Proceedings of the 41st International Conference on Machine
  Learning}, volume 235 of \emph{Proceedings of Machine Learning Research},
  pages 35104--35137. PMLR, 2024.
\newblock URL \url{https://proceedings.mlr.press/v235/matthews24a.html}.

\bibitem[Paglieri et~al.(2024)Paglieri, Cupia{\l}, Coward, Piterbarg, Wolczyk,
  Khan, Pignatelli, Kuci{\'n}ski, Pinto, Fergus, Foerster, Parker-Holder, and
  Rockt{\"a}schel]{paglieri2024balrog}
Davide Paglieri, Bart{\l}omiej Cupia{\l}, Samuel Coward, Ulyana Piterbarg,
  Maciej Wolczyk, Akbir Khan, Eduardo Pignatelli, {\L}ukasz Kuci{\'n}ski,
  Lerrel Pinto, Rob Fergus, Jakob~Nicolaus Foerster, Jack Parker-Holder, and
  Tim Rockt{\"a}schel.
\newblock {BALROG}: Benchmarking agentic {LLM} and {VLM} reasoning on games.
\newblock \emph{arXiv preprint arXiv:2411.13543}, 2024.
\newblock URL \url{https://arxiv.org/abs/2411.13543}.

\bibitem[Shinn et~al.(2023)Shinn, Cassano, Berman, Gopinath, Narasimhan, and
  Yao]{shinn2023reflexion}
Noah Shinn, Federico Cassano, Edward Berman, Ashwin Gopinath, Karthik
  Narasimhan, and Shunyu Yao.
\newblock {Reflexion}: Language agents with verbal reinforcement learning.
\newblock \emph{arXiv preprint arXiv:2303.11366}, 2023.
\newblock URL \url{https://arxiv.org/abs/2303.11366}.

\bibitem[Shu et~al.(2026)Shu, Zhang, Cho, Yang, Yuan, Zheng, Guntuku, Ungar,
  Yu, and Zhang]{shu2026agentworld}
Raphael Shu, Yusen Zhang, Young~Min Cho, Jin~Mo Yang, Yuan Yuan, Wenliang
  Zheng, Sharath~Chandra Guntuku, Lyle Ungar, Zhou Yu, and Rui Zhang.
\newblock {AgentWorld}: Benchmarking long-horizon collaboration of multi-agent
  {LLMs}.
\newblock \emph{arXiv preprint arXiv:2609.31590}, 2026.
\newblock URL \url{https://arxiv.org/abs/2609.31590}.

\bibitem[Suarez et~al.(2019)Suarez, Du, Isola, and
  Mordatch]{suarez2019neuralmmo}
Joseph Suarez, Yilun Du, Phillip Isola, and Igor Mordatch.
\newblock {Neural MMO}: A massively multiagent game environment for training
  and evaluating intelligent agents.
\newblock \emph{arXiv preprint arXiv:1903.00784}, 2019.
\newblock URL \url{https://arxiv.org/abs/1903.00784}.

\bibitem[Wang et~al.(2023)Wang, Xie, Jiang, Mandlekar, Xiao, Zhu, Fan, and
  Anandkumar]{wang2023voyager}
Guanzhi Wang, Yuqi Xie, Yunfan Jiang, Ajay Mandlekar, Chaowei Xiao, Yuke Zhu,
  Linxi Fan, and Anima Anandkumar.
\newblock {Voyager}: An open-ended embodied agent with large language models.
\newblock \emph{arXiv preprint arXiv:2305.16291}, 2023.
\newblock URL \url{https://arxiv.org/abs/2305.16291}.

\bibitem[Wang et~al.(2022)Wang, Jansen, C{\^o}t{\'e}, and
  Ammanabrolu]{wang2022scienceworld}
Ruoyao Wang, Peter Jansen, Marc-Alexandre C{\^o}t{\'e}, and Prithviraj
  Ammanabrolu.
\newblock {ScienceWorld}: Is your agent smarter than a 5th grader?
\newblock \emph{arXiv preprint arXiv:2203.07540}, 2022.
\newblock URL \url{https://arxiv.org/abs/2203.07540}.

\bibitem[Yao et~al.(2022)Yao, Zhao, Yu, Du, Shafran, Narasimhan, and
  Cao]{yao2022react}
Shunyu Yao, Jeffrey Zhao, Dian Yu, Nan Du, Izhak Shafran, Karthik Narasimhan,
  and Yuan Cao.
\newblock {ReAct}: Synergizing reasoning and acting in language models.
\newblock \emph{arXiv preprint arXiv:2210.03629}, 2022.
\newblock URL \url{https://arxiv.org/abs/2210.03629}.

\bibitem[Yao et~al.(2024)Yao, Shinn, Razavi, and Narasimhan]{yao2024taubench}
Shunyu Yao, Noah Shinn, Pedram Razavi, and Karthik Narasimhan.
\newblock {\(\tau\)}-bench: A benchmark for tool-agent-user interaction in
  real-world domains.
\newblock \emph{arXiv preprint arXiv:2406.12045}, 2024.
\newblock URL \url{https://arxiv.org/abs/2406.12045}.

\bibitem[Yu et~al.(2024)Yu, Fu, Deng, and Han]{yu2024mineland}
Xianhao Yu, Jiaqi Fu, Renjia Deng, and Wenjuan Han.
\newblock {MineLand}: Simulating large-scale multi-agent interactions with
  limited multimodal senses and physical needs.
\newblock \emph{arXiv preprint arXiv:2403.19267}, 2024.
\newblock URL \url{https://arxiv.org/abs/2403.19267}.

\bibitem[Zhou et~al.(2023)Zhou, Xu, Zhu, Zhou, Lo, Sridhar, Cheng, Ou, Bisk,
  Fried, Alon, and Neubig]{zhou2023webarena}
Shuyan Zhou, Frank~F. Xu, Hao Zhu, Xuhui Zhou, Robert Lo, Abishek Sridhar,
  Xianyi Cheng, Tianyue Ou, Yonatan Bisk, Daniel Fried, Uri Alon, and Graham
  Neubig.
\newblock {WebArena}: A realistic web environment for building autonomous
  agents.
\newblock \emph{arXiv preprint arXiv:2307.13854}, 2023.
\newblock URL \url{https://arxiv.org/abs/2307.13854}.

\end{thebibliography}

\endgroup
\clearpage
\appendix
\section{Additional Experimental Details}\label{sec:configuration}
All evaluated configurations use the same general-purpose coding harness, Claude Code, in an isolated container environment with reasoning effort set to high. Context is compacted automatically using a unified compaction window across configurations to prevent silent context truncation over the seven-hour run. Configuration names identify the foundation model served to the harness; exact upstream provider weights and minor snapshot revisions are controlled by the serving endpoints.

In the main evaluation, all characters are registered in a newly generated world and released simultaneously via a host synchronization barrier. The world contains 9 expedition regions (five three-floor dungeons, a single-floor sewer dungeon, two single-floor indoor service regions, and a five-floor sky tower totaling 23 floors), 8 regional bosses, 15 craftable tools and equipment items spanning tiers T1--T3 across 5 equipment slots, 13 NPCs, 11 food varieties (including 5 cooked meal types), a 10-course onboarding curriculum, and a 37-component, four-stage shelter blueprint. Authoritative data sources comprise: (1) frozen database tables of command receipts, runtime actor facts, combat ranking records (\texttt{runtime\_ranking\_facts}), construction projects and pieces, party groups, and trading offers; (2) continuous position and scene-transition trajectories; and (3) host-side character state snapshots recorded every five minutes, supplemented by server-returned state responses logged in the harness execution trajectories.

\section{Cross-Dimensional Matrix and Progression Score Weights}\label{sec:score-weights}
Table~\ref{tab:curriculum} summarizes the resulting cross-dimensional evaluation matrix. Each cell maps one agent capability within one gameplay domain to a verifiable in-game task normalized by its world ceiling or evaluated as a binary condition. For conditional diagnostic cells such as error-driven learning and safe extraction, we distinguish three states: attained ($\checkmark$), condition triggered but unattained ($\times$), and prerequisite unentered (---) when an agent never starves, never suffers a combat setback, never attempts building, or never triggers a material-shortage error.

\begin{table}[htbp]
\centering\small
\setlength{\tabcolsep}{4pt}
\caption{Cross-dimensional evaluation matrix intersecting the four gameplay domains with six core agent capabilities. Ceilings in parentheses reflect the fixed world configuration.}\label{tab:curriculum}
\begin{tabularx}{\linewidth}{@{}lYYYY@{}}\toprule
Capability & Survival & Combat & Building & Crafting \\\midrule
\textbf{C} Cognition & Food variety (/11) & Exploration (/9 regions) & Site selection (valid/tried) & Curriculum (/10 courses) \\\addlinespace[2pt]
\textbf{P} Planning & Farming loop (/3 steps) & Dungeon depth (/5 floors) & Blueprinting (/1 project) & Gear tier (/T3) \\\addlinespace[2pt]
\textbf{E} Execution & Cooked meals (/5 types) & Bosses defeated (/8) & Construction (/37 pieces) & Tool variety (/15 items) \\\addlinespace[2pt]
\textbf{M} Self-Management & Homeostasis (\% of time) & Extraction (/entered) & Home security (yes/no) & Equipped slots (/5 slots) \\\addlinespace[2pt]
\textbf{L} Learning & Resilience (yes/no) & Adaptation (yes/no) & Relocation (yes/no) & Restock \& craft (yes/no) \\\addlinespace[2pt]
\textbf{S} Social & Food trade (yes/no) & Co-op expedition (yes/no) & Property trade (yes/no) & NPC engagement (/13 NPCs) \\\bottomrule
\end{tabularx}
\end{table}

\paragraph{Within-domain stage weighting.}
To summarize progression and rank configurations, we adopt a two-level weighting scheme motivated by prerequisite depth and world-package content volume. Within each gameplay domain $d$, individual content terms $x_{m,d,k} \in [0, 1]$ represent the attained fraction of the fixed world ceiling and are weighted by their progression stage---assigning weight $1$ to entry-level discovery, $2$ to mid-game sustained loops, and $3$ to late-stage mastery---so that early foraging or tutorial steps cannot outweigh endgame accomplishments:
\begin{equation}\label{eq:score}
P_{m,d} = 100 \times \frac{\sum_{k \in \mathcal{T}_d} w_{d,k}\, x_{m,d,k}}{\sum_{k \in \mathcal{T}_d} w_{d,k}}, \qquad x_{m,d,k} \in [0, 1].
\end{equation}
Under this progression structure, survival weights initial food discovery at $1$, the farming loop and physiological homeostasis uptime at $2$, and cooked meal diversity at $3$. Combat weights discovered regions at $1$, cumulative cleared floors and defeated monster species at $2$, and defeated regional bosses at $3$. Building weights the four shelter construction stages progressively from $1$ for the foundation, $2$ for walls and roofing, to $3$ for indoor furnishings, alongside $3$ for door-lock installation. Crafting weights onboarding courses at $1$, tool variety and equipped slots at $2$, and highest gear tier and NPC engagement at $3$.

\paragraph{Cross-domain composite score.}
To order configurations in the main leaderboard, we combine the four domain progression scores in proportion to their content breadth and dependency depth in the world package:
\begin{equation}\label{eq:composite}
S_m = 0.20\, P_{m,\text{surv}} + 0.35\, P_{m,\text{comb}} + 0.15\, P_{m,\text{build}} + 0.30\, P_{m,\text{craft}}.
\end{equation}
Combat and crafting carry the largest weights of $0.35$ and $0.30$ because they span multi-region expeditions and the three-tier technology tree that require continuous progression across the season. Survival receives a weight of $0.20$ for baseline physiological maintenance, while building receives $0.15$ because constructing a shelter is a bounded single-project task that saturates once its 37 components and door lock are completed. Table~\ref{tab:score-components} lists all within-domain and cross-domain weights. When two configurations tie on rounded composite score $S_m$, ties are broken by bosses defeated, deepest floor reached, and distinct tools crafted.

\begin{table}[htbp]
\centering\small
\setlength{\tabcolsep}{4.5pt}
\caption{Two-level weighting structure for domain progression scores $P_{m,d}$ (Equation~\eqref{eq:score}) and the composite leaderboard score $S_m$ (Equation~\eqref{eq:composite}). Each term $x_{m,d,k} \in [0,1]$ is normalized by its fixed world-package ceiling.}\label{tab:score-components}
\begin{tabularx}{\linewidth}{@{}llYrr@{}}\toprule
Domain ($W_d$) & Stage & Content Term $x_{m,d,k}$ & Ceiling & Stage Weight $w_{d,k}$ \\\midrule
\textbf{Survival} ($20\%$) & Early & Food varieties consumed or held & 11 types & $1\times$ \\
 & Mid & Three-step agriculture loop (sow, harvest, cook) & 3 steps & $2\times$ \\
 & Mid & Physiological homeostasis uptime (non-depleted snapshots) & $100\%$ & $2\times$ \\
 & Late & Cooked meal varieties produced & 5 types & $3\times$ \\\midrule
\textbf{Combat} ($35\%$) & Early & Expedition regions discovered & 9 regions & $1\times$ \\
 & Mid & Cumulative dungeon and tower floors cleared & 23 floors & $2\times$ \\
 & Mid & Distinct non-boss monster species defeated & 10 species & $2\times$ \\
 & Late & Regional bosses defeated & 8 bosses & $3\times$ \\\midrule
\textbf{Building} ($15\%$) & Early--Late & Four-stage shelter construction (foundation $1\times$, walls/doors $2\times$, roof $2\times$, indoor furnishings $3\times$) & 37 pieces & $8\times$ \\
 & Late & Door lock installed on completed shelter & Binary (0/1) & $3\times$ \\\midrule
\textbf{Crafting} ($30\%$) & Early & Onboarding curriculum courses completed & 10 courses & $1\times$ \\
 & Mid & Distinct tools and equipment items acquired & 15 items & $2\times$ \\
 & Mid & Equipment slots populated on character & 5 slots & $2\times$ \\
 & Late & Highest equipment tier reached & Tier 3 & $3\times$ \\
 & Late & Fixed NPCs engaged for commissions and trade & 13 NPCs & $3\times$ \\
\bottomrule\end{tabularx}
\end{table}

\section{Assessment Details and Unentered Prerequisites}\label{sec:reconstruction}
All 24 capability cells in Table~\ref{tab:curriculum} and the four stage-weighted domain progression scores in Equation~\eqref{eq:score} are computed deterministically from the authoritative server records, trajectories, and state snapshots without manual annotation.

For conditional diagnostic cells, distinguishing between an unentered prerequisite (``---'') and an unattained triggered condition ($\times$) prevents conflating domain selection with capability failure:
\begin{itemize}
\item \textbf{Survival Learning (Resilience):} Evaluated if the character's satiety depletes to $\le 1$ in either the five-minute host snapshots or the server-returned state responses recorded in the harness trajectory (taking the union of both authoritative sources so brief depletions between five-minute intervals are captured); attained ($\checkmark$) if satiety subsequently recovers to $\ge 30$ through autonomous foraging and eating. Configurations that never deplete satiety in either source (\texttt{gpt-6-astra}) are marked ``---'', whereas configurations that experience a brief depletion between five-minute snapshots and immediately recover (such as \texttt{claude-opus-5-5}) retain $100\%$ snapshot homeostasis while attaining $\checkmark$ in resilience.
\item \textbf{Combat Execution (Bosses), Self-Management (Extraction), and Learning (Adaptation):} Defeated monsters and bosses are verified from combat ranking records and encounter receipts. Extraction measures the ratio of entered expedition regions from which the character exits alive back to the overworld, determined from scene transitions in the position trajectory, and is marked ``---'' if the agent never enters a dungeon. Adaptation is evaluated only if the agent experiences at least one combat setback (an encounter ending in withdrawal or defeat); it is attained ($\checkmark$) if the agent wins a subsequent encounter at a later timestamp.
\item \textbf{Building Capabilities:} All building capability cells are marked ``---'' if the agent never issues a construction planning, siting, or assembly command. Relocation (\textbf{L}) is evaluated only if the agent encounters at least one invalid-site rejection during construction siting; it records whether a subsequent siting command succeeds, without penalizing the number of intermediate retries.
\item \textbf{Crafting Learning (Restock \& craft):} Evaluated only if the agent receives a material-shortage rejection when attempting a crafting command; attained ($\checkmark$) if a subsequent crafting command succeeds after replenishing materials.
\end{itemize}
\FloatBarrier

\end{document}